\documentclass{article}
\usepackage{spconf,amsmath,graphicx}
\usepackage{multirow}
\usepackage{amsfonts}
\usepackage{bm}
\usepackage{subcaption}

\DeclareMathOperator*{\argmax}{arg\,max}

\title{Diarisation using Location tracking with agglomerative clustering}
\name{Jeremy H. M. Wong, Igor Abramovski, Xiong Xiao, and Yifan Gong}
\address{Microsoft, USA}
\begin{document}
\ninept
\maketitle
\begin{abstract}

Previous works have shown that spatial location information can be complementary to speaker embeddings for a speaker diarisation task. However, the models used often assume that speakers are fairly stationary throughout a meeting. This paper proposes to relax this assumption, by explicitly modelling the movements of speakers within an Agglomerative Hierarchical Clustering (AHC) diarisation framework. Kalman filters, which track the locations of speakers, are used to compute log-likelihood ratios that contribute to the cluster affinity computations for the AHC merging and stopping decisions. Experiments show that the proposed approach is able to yield improvements on a Microsoft rich meeting transcription task, compared to methods that do not use location information or that make stationarity assumptions.

\end{abstract}
\section{Introduction}

Speaker diarisation aims to cluster together segments of audio that are uttered by the same speaker. This is useful in a rich meeting transcription task, where both the identity of a speaker and the word being said need to be hypothesised. Spectral \cite{ning2006} and $k$-means \cite{shum2011} clustering can be used for diarisation, after first estimating the number of clusters, by for example, finding the maximum gap in a chosen statistic \cite{tibshirani2002,park2020}. Agglomerative Hierarchical Clustering (AHC) \cite{siegler1997,jin1997} instead jointly estimates the cluster assignments and number of clusters. The Hidden Markov Model (HMM) can also be used, either to compute merging scores within AHC \cite{ajmera2003}, or on its own after having estimated the number of clusters \cite{diez2019,landini2020}. These methods often rely solely on features in the form of speaker embeddings, such as $i$ \cite{dehak2011}, $x$ \cite{snyder2018}, and $d$-vectors \cite{variani2014}. The speaker embeddings are intended to express information that is useful in discriminating between different speakers.

When multi-channel audio is available, it is possible to estimate the instantaneous location from where the sound originated from. This information may be complementary to the speaker embeddings in the diarisation task. Previous works have investigated using time-delay-of-arrival \cite{pardo2007,vijayasenan2012} and Sound Source Localisation (SSL) information \cite{wong2021}, together with speaker embeddings in HMM clustering. There are also a diversity of methods to count and localise multiple speakers, without using speaker embeddings \cite{murase2005,mcdonough2013,plinge2014}.

Speakers may move over the duration of a meeting. Explicitly modelling this movement may aid in diarisation. In multi-face tracking, Kalman filters are often used to track face movements from visual information \cite{shaik2007,foytik2011}. When multi-channel audio is available, acoustic location information has been shown to be complementary to visual information for face movement tracking \cite{gebru2015}. The LOCATA challenge \cite{evers2000} has helped to spur the development of audio-only location tracking methods. Several of these approaches also rely on Kalman filters, to track the locations of a single \cite{bechler2003,salvati2018} or multiple \cite{segura2008} audio sources.

This paper proposes to perform diarisation, while modelling the movements of multiple speakers. It builds upon the works in \cite{pardo2007,vijayasenan2012,wong2021}, by tracking the movements of speakers, rather than assuming that speakers are stationary. It also extends upon the audio-only tracking methods, such as in \cite{murase2005,mcdonough2013,plinge2014,segura2008}, by using both location information and speaker embeddings in diarisation. Diarisation is performed using AHC. Speaker movement is modelled as the likelihood of a sequence of instantaneous locations, computed using a Kalman filter. This is used together with a speaker embedding affinity score in the AHC cluster merging and stopping criteria.

\section{von Mises Kalman filter tracking}

The Kalman filter \cite{kalman1960} can be used to model movement through location tracking. Using the Markov assumptions, the Kalman filter computes the likelihood of an observation sequence as
\begin{equation}
p\left(\mathbf{X}_{1:T}\right)\approx\!\!\int\! p\left(z_1\right)p\left(\mathbf{x}_1\middle|z_1\right)\prod_{t=2}^Tp\left(z_t\middle|z_{t-1}\right)p\left(\mathbf{x}_t\middle|z_t\right)d\mathbf{z}_{1:T},
\label{eq:observation_sequence_likelihood}
\end{equation}
where $t$ is the frame index, $T$ is the total number of frames, and $\mathbf{x}_t$ is an observed instantaneous location feature, whose possible forms are discussed in section \ref{sec:observation_feature}. In this paper, the hidden state, $z_t$, represents the estimated location of the speaker. In the future, it may be beneficial to also investigate modelling the velocity and higher time derivatives in the hidden state, as is often done in face tracking \cite{shaik2007,foytik2011}.

In this paper, the speaker location is expressed as the horizontal angle around a microphone array. This is a continuous variable that is bounded within $\left(-\pi ,\pi\right]$ in radians, with a periodic boundary condition. Previous works have satisfied these properties using von Mises and warped normal density functions \cite{kurz2016}. The transition likelihood used in this paper is a von Mises density function,
\begin{equation}
p\left(z_t\middle|z_{t-1}\right)=\frac{1}{2\pi I_0\left(\kappa^z\right)}e^{\kappa^z\cos\left(z_t-z_{t-1}\right)},
\label{eq:transition}
\end{equation}
where $I_\nu$ is the modified Bessel function of the first kind of order $\nu$ and the concentration parameter, $\kappa^z$, expresses how tightly the density function is concentrated about the mean of $z_{t-1}$. A higher concentration yields a lower likelihood for protean sequences. The initial state likelihood, $p\left(z_1\right)$, is set to a uniform density function, as there is complete uncertainty of where the speaker is before any observation is made.

\subsection{Observation feature}
\label{sec:observation_feature}

Two forms of observed features are considered, namely the scalar Direction-Of-Arrival (DOA) and the full SSL vector. The SSL vector, $\mathbf{s}_t$, is a categorical distribution, with each dimension representing the probability that the sound had originated from the respective angular bin around the microphone array,
\begin{equation}
s_{ti}=P\left(\theta=i\middle|\mathbf{x}_t\right),
\end{equation}
where $i$ is the angular bin index and $\theta$ is the angular bin from which the audio forming feature $\mathbf{x}_t$ may have originated. The SSL is computed using a complex angular central Gaussian model \cite{ito2016}, as is described in \cite{yoshioka2019b}. The DOA, $\phi_t$, is computed as the mode of the SSL,
\begin{equation}
\phi_t=b_j\quad,\text{ where}\quad j=\argmax_is_{ti},
\end{equation}
and $b_j$ is the angle in radians of the $j$th bin. Instead of the mode, it is also possible to estimate the DOA as the circular mean of the SSL, which can be computed using \eqref{eq:circular_mean}. However, initial tests did not suggest any significant difference between the performances of either form of DOA.

When using the DOA as the observed feature, the Kalman filter is used to compute $p\left(\bm{\phi}_{1:T}\right)$, by substituting the placeholder, $\mathbf{x}_t$, with $\phi_t$ in \eqref{eq:observation_sequence_likelihood}. The emission likelihood is chosen to be a von Mises density function,
\begin{equation}
p\left(\phi_t\middle|z_t\right)=\frac{1}{2\pi I_0\left(\kappa^\phi\right)}e^{\kappa^\phi\cos\left(\phi_t-z_t\right)},
\label{eq:doa_emission}
\end{equation}
where the concentration parameter, $\kappa^\phi$, expresses the random noise in the observation.

The DOA only represents an estimate of the instantaneous angle of the speaker. However, interactions with the environment, noise, and the limited spatial resolution of the microphone array geometry may result in uncertainty in this estimation. The full SSL vector may express information about this uncertainty. The likelihood of an SSL sequence, $p\left(\mathbf{S}_{1:T}\right)$, can be computed by substituting $\mathbf{x}_t$ with $\mathbf{s}_t$ in \eqref{eq:observation_sequence_likelihood}, and using an emission likelihood in the form of a continuous categorical density function \cite{gordonrodriguez2020},
\begin{equation}
p\left(\mathbf{s}_t\middle|z_t\right)=\frac{1}{C\left(\bm{\lambda}_t\right)}\prod_{i=1}^N\lambda_{ti}^{s_{ti}},
\label{eq:continuous_categorical}
\end{equation}
where $N$ is the number of angular bins, $C\left(\bm{\lambda}_t\right)$ is the normalisation constant defined in \cite{gordonrodriguez2020}, and the continuous categorical bin probabilities are computed as a discretised von Mises distribution about the mean of $z_t$,
\begin{equation}
\lambda_{ti}=\frac{e^{\kappa^\phi\cos\left(b_i-z_t\right)}}{\sum\limits_{j=1}^Ne^{\kappa^\phi\cos\left(b_j-z_t\right)}}.
\label{eq:discrete_vm}
\end{equation}
The continuous categorical emission likelihood can be interpreted as being that if multiple samples are drawn when given the same $z_t$, then \eqref{eq:continuous_categorical} computes the likelihood of observing each angular bin, $i$, at a fraction of $s_{ti}$, out of all of the samples. If a Dirichlet density function is used instead, by swapping the places of $\lambda_{ti}$ and $s_{ti}$ in \eqref{eq:continuous_categorical}, then the interpretation will be different, and the simplification of \eqref{eq:equivalent_vm} will no longer be applicable. By taking the logarithm of \eqref{eq:continuous_categorical}, the emission log-likelihood can be seen to be a Kullback Leibler (KL)-divergence between two categorical distributions, one of the observation, $\mathbf{s}_t$, and the other being a prediction of the angular distribution, $\bm{\lambda}_t$. That is to say, that when the model predicts the angular state to be $z_t$, the model also predicts that the observed SSL should be similar to $\bm{\lambda_t}$. The emission log-likelihood then measures a similarity score between the observed SSL, $\mathbf{s}_t$, and the predicted SSL, $\bm{\lambda}_t$. The work in \cite{wong2021} also uses an emission log-likelihood in the form of a KL-divergence between SSL vectors, for HMM diarisation.

However, the exact form of \eqref{eq:continuous_categorical} presents a challenge, as the normalisation term of $C\left(\bm{\lambda}_t\right)$ is difficult to compute in a numerically stable manner \cite{gordonrodriguez2020}. The approximation is therefore made that $C\left(\bm{\lambda}_t\right)$ is independent of $z_t$, thereby allowing $C\left(\bm{\lambda}_t\right)$ to be ignored when computing log-likelihood ratios for the AHC affinity scores, as will be described in Section \ref{sec:ahc}. By ignoring $C\left(\bm{\lambda}_t\right)$ and substituting in \eqref{eq:discrete_vm}, \eqref{eq:continuous_categorical} can be simplified to
\begin{align}
p\left(\mathbf{s}_t\middle|z_t\right)&\propto\prod_{i=1}^N\lambda_{ti}^{s_{ti}}\notag\\
&=\frac{e^{\rho_t\cos\left(\mu_t-z_t\right)}}{\sum\limits_{j=1}^Ne^{\kappa^\phi\cos\left(b_j-z_t\right)}},\label{eq:equivalent_vm}
\end{align}
where
\begin{equation}
\rho_t=\kappa^\phi\sqrt{\sum_{i=1}^N\sum_{j=1}^Ns_{ti}s_{tj}\cos\left(b_i-b_j\right)}
\label{eq:ssl_concentration}
\end{equation}
and
\begin{equation}
\mu_t=\tan^{-1}\left(\frac{\sum\limits_{i=1}^Ns_{ti}\sin b_i}{\sum\limits_{i=1}^Ns_{ti}\cos b_i}\right).
\label{eq:circular_mean}
\end{equation}
The form of \eqref{eq:equivalent_vm} is reminiscent of a von Mises density function. Therefore, by choosing to use a combination of a continuous categorical density function and a discretised von Mises distribution for the SSL emission likelihood in \eqref{eq:continuous_categorical} and \eqref{eq:discrete_vm}, the effective emission likelihood will also look similar to a von Mises density function, and $\mathbf{s}_t$ at each frame can be completely summarised by its equivalent concentration, $\rho_t$, and circular mean, $\mu_t$. The concentration, $\rho_t$, may weigh the contribution of each frame to the total log-likelihood proportionally to the sharpness of the SSL distribution.

\begin{figure}[t]
\centering
\begin{subfigure}[b]{0.239\textwidth}
\centering
\includegraphics[width=\textwidth]{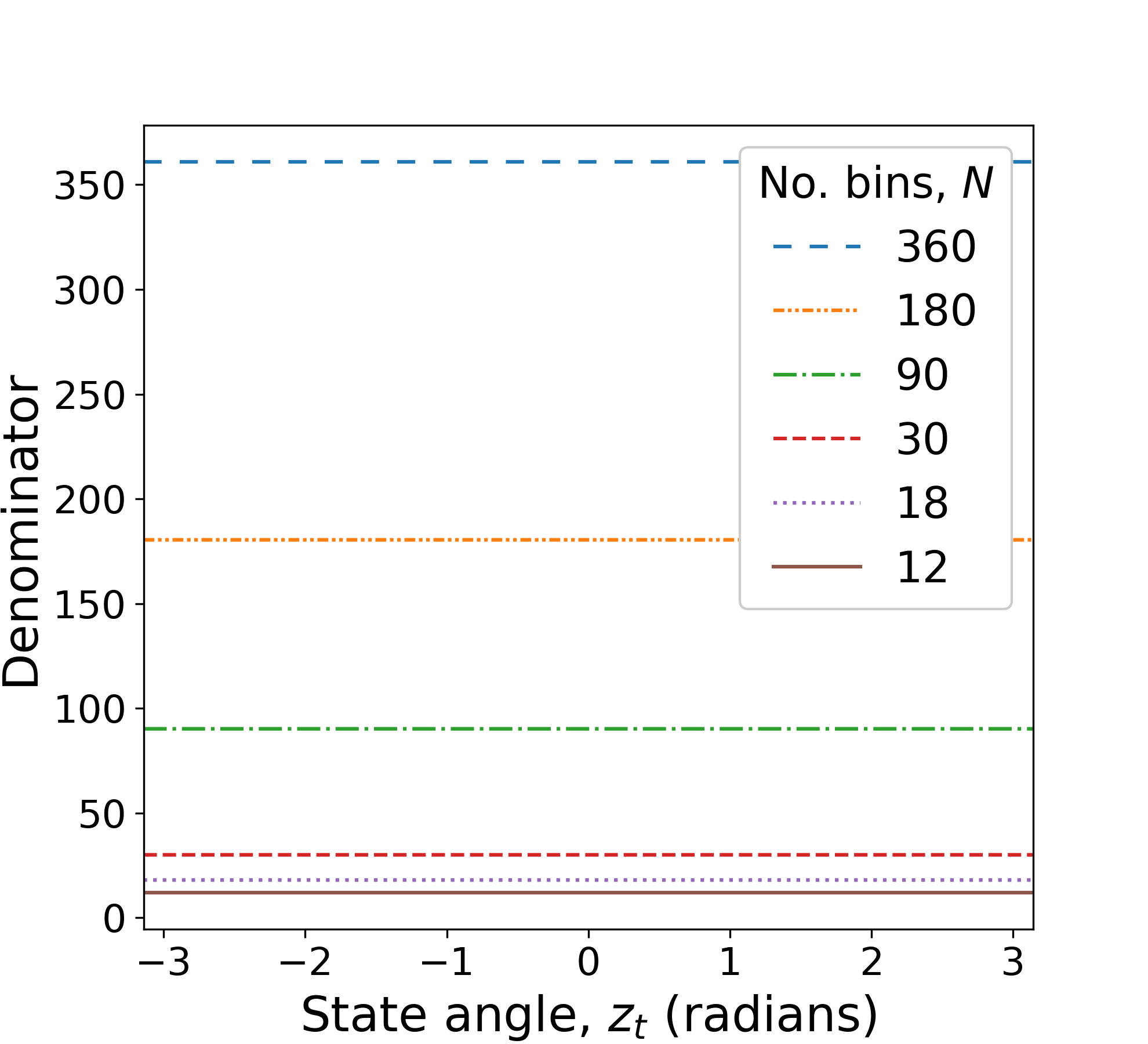}
\caption{$\kappa^\phi =0.1$}
\end{subfigure}
\begin{subfigure}[b]{0.239\textwidth}
\centering
\includegraphics[width=\textwidth]{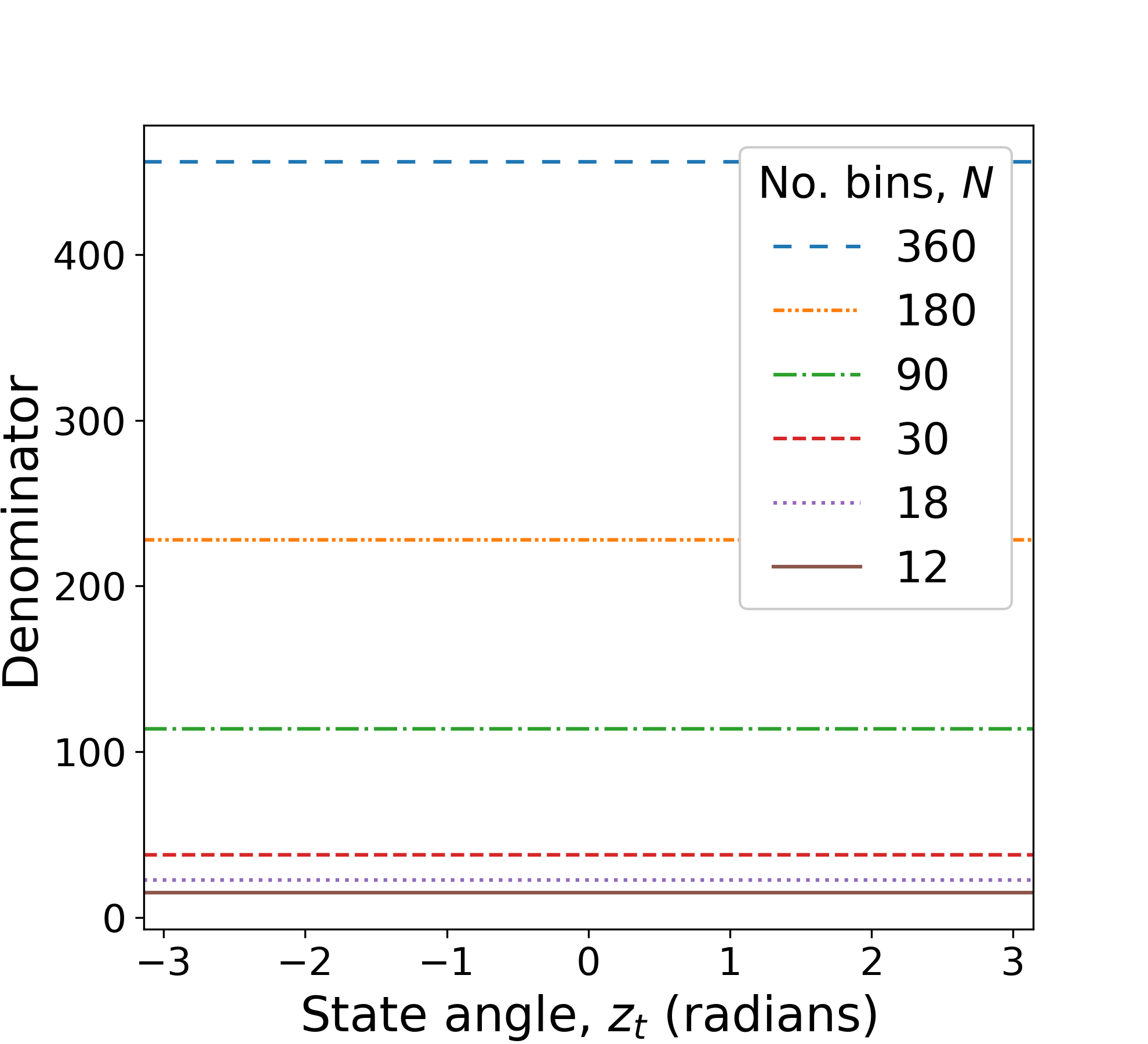}
\caption{$\kappa^\phi =1$}
\end{subfigure}
\begin{subfigure}[b]{0.239\textwidth}
\centering
\includegraphics[width=\textwidth]{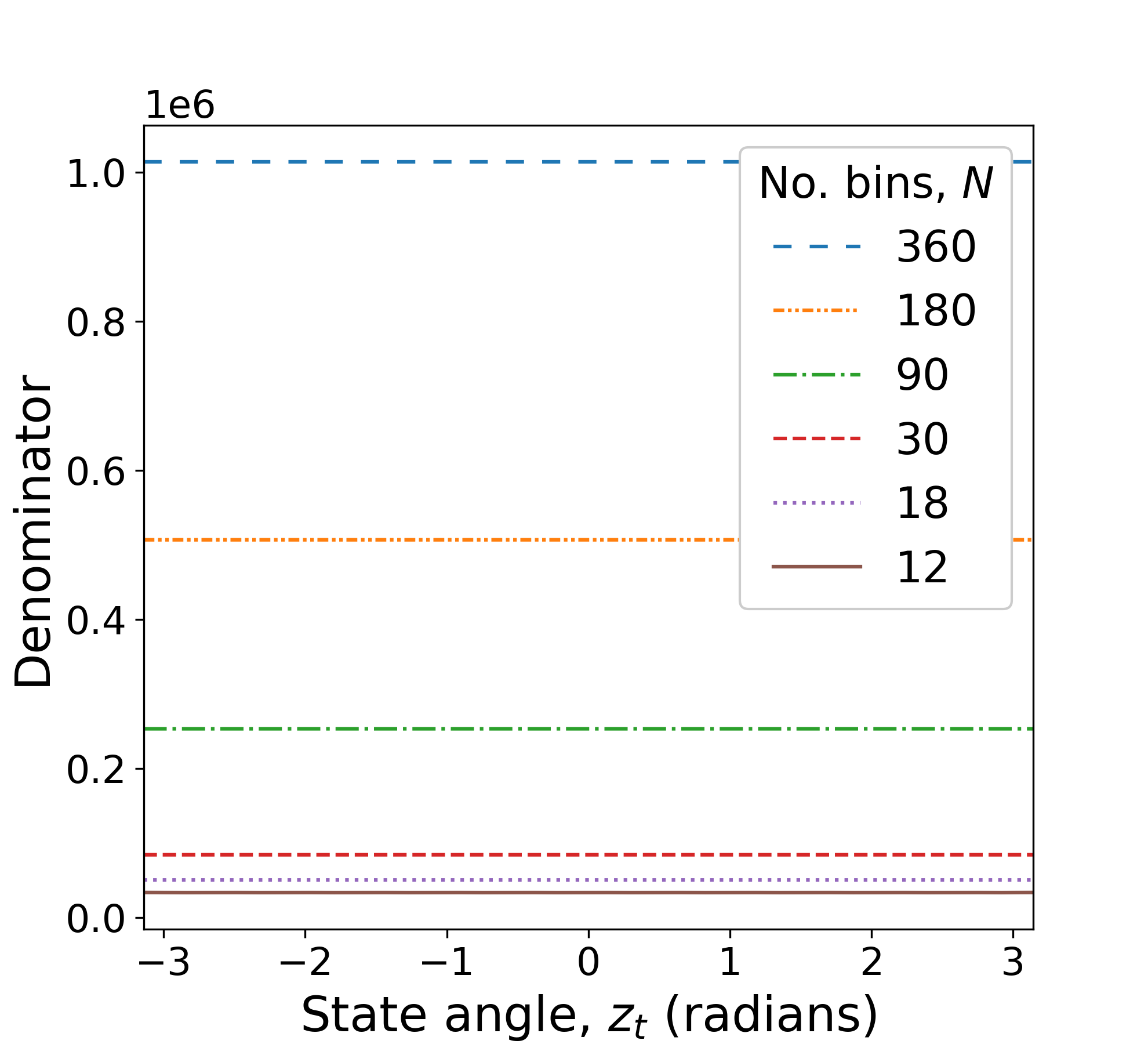}
\caption{$\kappa^\phi =10$}
\end{subfigure}
\begin{subfigure}[b]{0.239\textwidth}
\centering
\includegraphics[width=\textwidth]{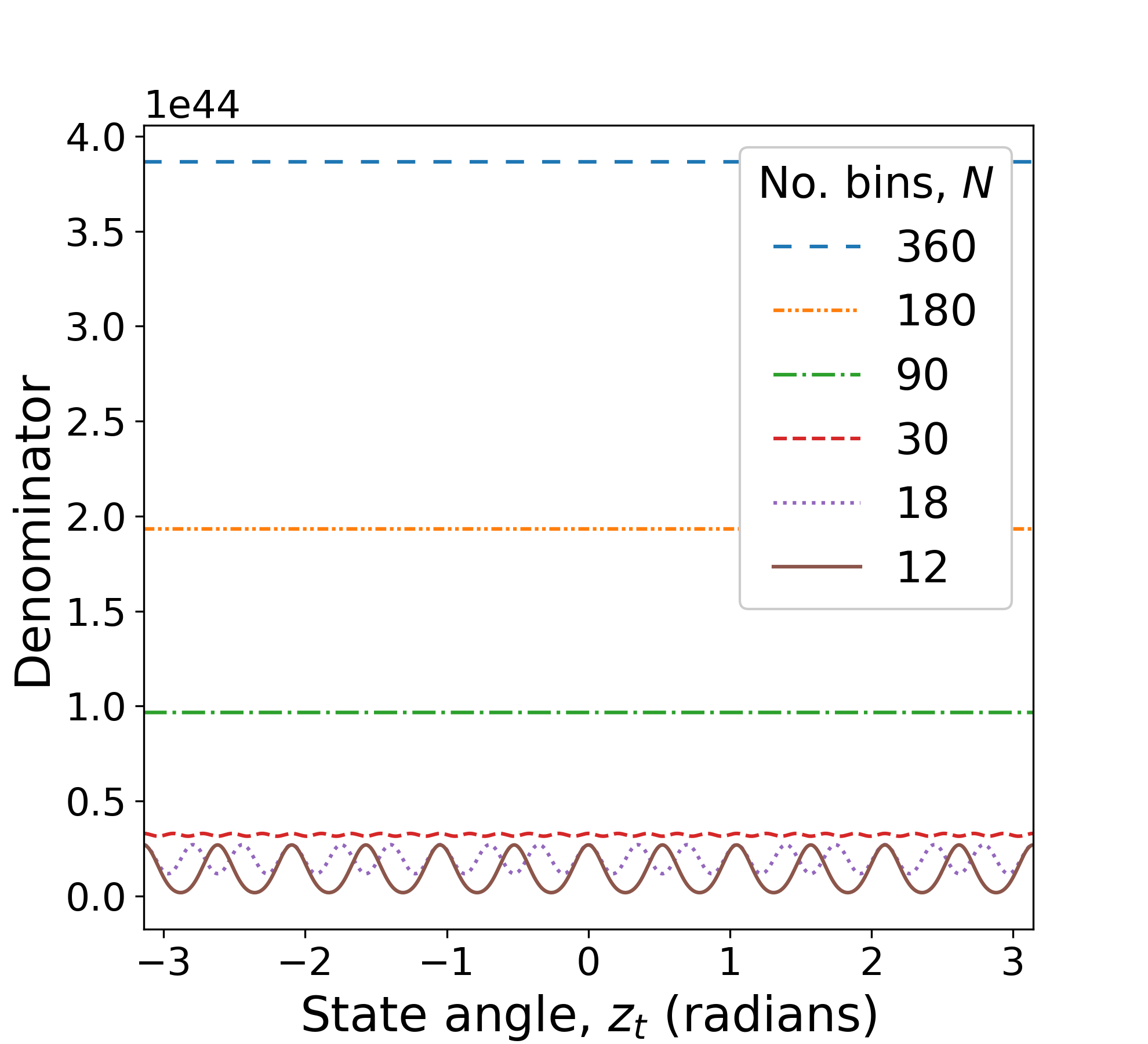}
\caption{$\kappa^\phi =100$}
\end{subfigure}
\caption{Denominator term of discretised von Mises distribution \eqref{eq:discrete_vm}}
\label{fig:norm_const}
\end{figure}

However, unlike a von Mises density function, the denominator in \eqref{eq:discrete_vm} and \eqref{eq:equivalent_vm} depends on $z_t$, which is inconvenient for the forward recursion integral described in Section \ref{sec:likelihood_computation}. Figure \ref{fig:norm_const} plots the denominator, $\sum_je^{\kappa^\phi\cos\left(b_j-z_t\right)}$, over a variety of $\kappa^\phi$ and $N$ values. It can be seen that the denominator is approximately independent of $z_t$, except when the concentration, $\kappa^\phi$, is large at the same time as the number of angular bins, $N$, is small. The experiments in this paper do not operate in such a regime. Therefore, it may be reasonable to approximate the denominator as being independent of $z_t$. This allows the SSL emission likelihood to be expressed as
\begin{equation}
p\left(\mathbf{s}_t\middle|z_t\right)\propto e^{\rho_t\cos\left(\mu_t-z_t\right)},
\label{eq:ssl_emission}
\end{equation}
which has the form of a von Mises density function over $z_t$.

\subsection{Likelihood computation}
\label{sec:likelihood_computation}

The Kalman filter can be used to compute the log-likelihood of a DOA or SSL observation sequence, $\log p\left(\mathbf{X}_{1:T}\right)$. This log-likelihood can be used as a score in the AHC merging and stopping criteria, as will be discussed in Section \ref{sec:ahc}. The likelihood can be computed using the Kalman filter forward recursion,
\begin{equation}
p\left(z_t\middle|\mathbf{X}_{1:t}\right)\propto p\left(\mathbf{x}_t\middle|z_t\right)\int p\left(z_{t-1}\middle|\mathbf{X}_{1:t-1}\right)p\left(z_t\middle|z_{t-1}\right)dz_{t-1}.
\end{equation}
This can be broken down into the prediction step,
\begin{equation}
p\left(z_t\middle|\mathbf{X}_{1:t-1}\right)=\int p\left(z_{t-1}\middle|\mathbf{X}_{1:t-1}\right)p\left(z_t\middle|z_{t-1}\right)dz_{t-1}
\label{eq:prediction}
\end{equation}
and the update step,
\begin{equation}
p\left(z_t\middle|\mathbf{X}_{1:t}\right)\propto p\left(\mathbf{x}_t\middle|z_t\right)p\left(z_t\middle|\mathbf{X}_{1:t-1}\right).
\label{eq:update}
\end{equation}

In this paper, the transition likelihood in \eqref{eq:transition}, and emission likelihoods in \eqref{eq:doa_emission} and \eqref{eq:ssl_emission} all have the form of von Mises density functions in terms of the random variable $z_t$. The prediction step in \eqref{eq:prediction} is a convolution operation. Unfortunately, the von Mises density function is not closed under convolution, but instead the result takes a form described in \cite{mardia1999}. However, it has been shown that the result of the convolution can be closely approximated by a von Mises density function \cite{stephens1963}, thereby allowing \eqref{eq:prediction} to be expressed as
\begin{equation}
p\left(z_t\middle|\mathbf{X}_{1:t-1}\right)\approx\frac{1}{2\pi I_0\left(\lambda_t^\prime\right)}e^{\lambda_t^\prime\cos\left(z_t-\eta_t^\prime\right)}.
\label{eq:prediction_vm}
\end{equation}
The prediction concentration is
\begin{equation}
\lambda_t^\prime=A^{-1}\left[A\left(\lambda_{t-1}\right)A\left(\kappa^z\right)\right]
\label{eq:prediction_concentration}
\end{equation}
and the prediction mean is
\begin{equation}
\eta_t^\prime=\eta_{t-1},
\label{eq:prediction_mean}
\end{equation}
where $A^{-1}$ is the functional inverse of
\begin{equation}
A\left(\kappa\right)=\frac{I_1\left(\kappa\right)}{I_0\left(\kappa\right)},
\end{equation}
which can be solved for using the Newton-Raphson root finding algorithm, and both $\lambda_{t-1}$ and $\eta_{t-1}$ are the parameters of the update step von Mises density function from the previous frame. In the prediction step, the concentration is broadened from the previous frame, through the inclusion of $\kappa^z$ in \eqref{eq:prediction_concentration}. The mean in \eqref{eq:prediction_mean} does not change, because higher temporal derivatives of the angle are not modelled in the hidden state.

The von Mises density function is closed under multiplication. Therefore, by using the approximate prediction in \eqref{eq:prediction_vm}, the update step of \eqref{eq:update} becomes
\begin{equation}
p\left(z_t\middle|\mathbf{X}_{1:t}\right)=\frac{1}{2\pi I_0\left(\lambda_t\right)}e^{\lambda_t\cos\left(z_t-\eta_t\right)}.
\label{eq:update_vm}
\end{equation}
When using DOA observations with an emission likelihood of \eqref{eq:doa_emission}, the update concentration is
\begin{equation}
\lambda_t=\sqrt{\kappa^{\phi 2}+\lambda_t^{\prime 2}+2\kappa^\phi\lambda_t^\prime\cos\left(\phi_t-\eta_t^\prime\right)}
\end{equation}
and the update mean is
\begin{equation}
\eta_t=\tan^{-1}\left(\frac{\kappa^\phi\sin\phi_t+\lambda_t^\prime\sin\eta_t^\prime}{\kappa^\phi\cos\phi_t+\lambda_t^\prime\cos\eta_t^\prime}\right).
\label{eq:update_mean_doa}
\end{equation}
When instead using SSL observations with an emission likelihood of \eqref{eq:ssl_emission}, the update concentration is
\begin{equation}
\lambda_t=\sqrt{\rho_t^2+\lambda_t^{\prime 2}+2\rho_t\lambda_t^\prime\cos\left(\mu_t-\eta_t^\prime\right)}
\end{equation}
and the update mean is
\begin{equation}
\eta_t=\tan^{-1}\left(\frac{\rho_t\sin\mu_t+\lambda_t^\prime\sin\eta_t^\prime}{\rho_t\cos\mu_t+\lambda_t^\prime\cos\eta_t^\prime}\right).
\label{eq:update_mean_ssl}
\end{equation}
The mean updates of \eqref{eq:update_mean_doa} and \eqref{eq:update_mean_ssl} are weighted circular averages between the prediction mean and the observation. This serves to bring $\eta_t$ closer to the current observation.

After having computed the prediction density function through the forward recursion, the log-likelihood of the observation sequence can then be computed as
\begin{align}
\log p\left(\mathbf{X}_{1:T}\right)&=\sum_{t=1}^T\log p\left(\mathbf{x}_t\middle|\mathbf{X}_{1:t-1}\right)\notag\\
&=\sum_{t=1}^T\log\int p\left(\mathbf{x}_t\middle|z_t\right)p\left(z_t\middle|\mathbf{X}_{1:t-1}\right)dz_t.
\label{eq:conditional_loglikelihood}
\end{align}
When computing the log-likelihood, there is no need to preserve the von Mises form, as only a point estimate is needed. Therefore, the exact convolution \cite{mardia1999} can be used, which for DOA observations is
\begin{equation}
p\left(\phi_t\middle|\bm{\phi}_{1:t-1}\right)=\frac{I_0\left[\sqrt{\kappa^{\phi 2}+\lambda_t^{\prime 2}+2\kappa^\phi\lambda_t^\prime\cos\left(\phi_t-\eta_t^\prime\right)}\right]}{2\pi I_0\left(\kappa^\phi\right)I_0\left(\lambda_t^\prime\right)}.
\label{eq:exact_convolution}
\end{equation}
When using SSL observations, the exact emission likelihood of \eqref{eq:continuous_categorical} is difficult to use, because of the numerical instability of the normalisation term. Therefore, the experiments in this paper approximate the SSL observation sequence log-likelihood using the same form as \eqref{eq:exact_convolution}, with $\kappa^\phi$ and $\phi_t$ being substituted with $\rho_t$ and $\mu_t$ respectively. This approximation again ignores the denominator terms in both \eqref{eq:continuous_categorical} and \eqref{eq:discrete_vm}, and re-normalises \eqref{eq:ssl_emission} over $z_t$.

A speaker may have discontiguous regions of speech, since a speaker may not speak continuously throughout a whole meeting. As such, when modelling a speaker's movement with a Kalman filter, there may be frames for which the speaker has no DOA or SSL observation. This is analogous to an occlusion in the visual face tracking task. For such frames, $\mathbf{x}_t$ is undefined and the emission likelihood is simply set to $p\left(\mathbf{x}_t\middle|z_t\right)=1$. The update step for these frames simplifies from \eqref{eq:update} to,
\begin{equation}
p\left(z_t\middle|\mathbf{X}_{1:t}\right)= p\left(z_t\middle|\mathbf{X}_{1:t-1}\right).
\end{equation}
When computing the conditional log-likelihood for such frames, using $p\left(\mathbf{x}_t\middle|z_t\right)=1$ in \eqref{eq:conditional_loglikelihood} yields $\log p\left(\mathbf{x}_t\middle|\mathbf{X}_{1:t-1}\right)=0$. Thus these frames, without observations, do not contribute to the total observation sequence log-likelihood.

In the setup used in this paper, diarisation is performed after speech separation, to handle overlapped speech. As such, it is possible to encounter situations where the observation sequence may have frames from multiple separated channels that overlap in time. Let $\mathbf{x}_t^{\left(1\right)}$ and $\mathbf{x}_t^{\left(2\right)}$ be observations at frame index $t$ from channels 1 and 2 respectively. During the update step in \eqref{eq:update} and the likelihood computation in \eqref{eq:conditional_loglikelihood}, the emission likelihood of $p\left(\mathbf{x}_t\middle|z_t\right)$ is simply substituted with $p\left(\mathbf{x}_t^{\left(1\right)}\middle|z_t\right)p\left(\mathbf{x}_t^{\left(2\right)}\middle|z_t\right)$, where it is assumed that the parallel observations of $\mathbf{x}_t^{\left(1\right)}$ and $\mathbf{x}_t^{\left(2\right)}$ are conditionally independent of each other when given $z_t$. As a reminder, the product of multiple von Mises density functions also has the form of a von Mises density function.

\subsection{Parameter estimation}
\label{sec:parameter_estimation}

The Kalman filter has two parameters, $\kappa^z$ and $\kappa^\phi$. These express the dynamic ranges of the speaker's movement speed and random noise in the observation respectively. One possible method of estimating them is through maximising the log-likelihood of the observation sequence,
\begin{equation}
\kappa^{z*},\kappa^{\phi *}=\argmax_{\kappa^z,\kappa^\phi}\log p\left(\mathbf{X}_{1:T}\middle|\kappa^z,\kappa^\phi\right),
\label{eq:maximum_likelihood}
\end{equation}
using the Expectation-Maximisation (EM) algorithm.

The E-step requires the computation of the state posteriors,
\begin{equation}
p\left(z_t\middle|\mathbf{X}_{1:T}\right)\propto p\left(z_t\middle|\mathbf{X}_{1:t}\right)p\left(\mathbf{X}_{t+1:T}\middle|z_t\right),
\end{equation}
which is a product between the forward and backward density functions. The backward density function can be expressed as
\begin{equation}
p\left(\mathbf{X}_{t+1:T}\middle|z_t\right)\!\propto\!\!\!\int\!\! p\left(\mathbf{X}_{t+2:T}\middle|z_{t+1}\right)p\left(\mathbf{x}_{t+1}\middle|z_{t+1}\right)p\left(z_{t+1}\middle|z_t\right)dz_{t+1}\!.
\label{eq:backward}
\end{equation}
This can be computed using the product and approximate convolution of von Mises density functions, described in Section \ref{sec:likelihood_computation}. An analogous expression can also be derived for the joint state posterior, $p\left(z_t,z_{t+1}\middle|\mathbf{X}_{1:T}\right)$, which is omitted here for brevity. For frames that do not have an observation of the speaker's location, $p\left(\mathbf{x}_t\middle|z_t\right)=1$ in \eqref{eq:backward}.

The M-step update for $\kappa^\phi$ with DOA observations is
\begin{align}
\kappa_{\left(u+1\right)}^\phi=A^{-1}\left[\frac{1}{T}\sum_{t=1}^T\right.&\int p\left(z_t\middle|\mathbf{X}_{1:T},\kappa_{\left(u\right)}^\phi,\kappa_{\left(u\right)}^z\right)\notag\\
&\left.\vphantom{\sum_{t=1}^T}\times\cos\left(\phi_t-z_t\right)dz_t\right]
\label{eq:m_step_observation}
\end{align}
and that for $\kappa^z$ is
\begin{align}
\kappa_{\left(u+1\right)}^z=A^{-1}\left[\frac{1}{T-1}\sum_{t=1}^{T-1}\right.&\int p\left(z_t,z_{t+1}\middle|\mathbf{X}_{1:T},\kappa_{\left(u\right)}^\phi,\kappa_{\left(u\right)}^z\right)\notag\\
&\left.\vphantom{\sum_{t=1}^{T-1}}\times\cos\left(z_{t+1}-z_t\right)dz_tdz_{t+1}\right],
\label{eq:m_step_transition}
\end{align}
where $u$ is the EM iteration index. Monte Carlo approximations can be used to compute the integrals in \eqref{eq:m_step_observation} and \eqref{eq:m_step_transition}.

Estimating the parameters by maximising the likelihood only makes sense when using DOA observations. When using SSL observations, the normalising terms in \eqref{eq:continuous_categorical} and \eqref{eq:discrete_vm} are dependent on $\kappa^\phi$. Therefore a naive maximum likelihood optimisation with the previously described SSL observation approximations, of omitting the normalising terms, may not converge. In the experiments presented in this paper, $\kappa^\phi$ and $\kappa^z$ were optimised for DOA observations, and the same parameter values were then also used for SSL observations.

The approximate convolution used in the forward and backward recursions yields E-step posteriors and M-step updates that are also approximate. Therefore, there is no guarantee that the observation sequence log-likelihood will not worsen at each iteration. Furthermore, there is no guarantee that the parameters will converge to locally optimal values. However, the convolution approximation is needed to allow the density functions to remain closed within the von Mises family, and simplify the mathematics for the recursions. In the future, it may be interesting to investigate parameter estimation methods with fewer approximations.

\section{Agglomerative hierarchical clustering}
\label{sec:ahc}

Speaker diarisation can be performed using AHC. The aim is to cluster together segments that belong to the same speaker. AHC begins by treating each segment as a separate cluster. At each iteration, the two clusters with the highest affinity score are merged in a greedy manner. The merging iterations continue until the maximum remaining affinity falls below a threshold. Work in \cite{ajmera2003} uses the Bayes' Information Criterion (BIC) as the affinity and computes the observation sequence likelihoods for the clusters through a HMM with Gaussian mixture model emission likelihoods. The BIC allows the model complexity to remain constant through all AHC iterations, thereby alleviating any favouritism toward having more speakers. However, the stopping threshold for the BIC can be difficult to tune robustly in practice. Furthermore, each AHC merger iteration requires an optimisation of the HMM to be run using the EM algorithm, which can be computationally expensive.

The setup in this paper instead uses a simpler AHC formulation that follows \cite{wong2021}, where the affinity is computed as a cosine similarity between the speaker embedding centroid of each cluster. Given two clusters, $n$ and $m$, with unit-length speaker embeddings $\mathbf{d}_n$ and $\mathbf{d}_m$ respectively, the affinity is
\begin{equation}
\mathcal{A}_\text{speaker}\left(n,m\right)=\mathbf{d}_n\cdot\mathbf{d}_m.
\label{eq:affinity_dvec}
\end{equation}
This measures a similarity between the speaker embeddings of the two clusters. These embeddings are often extracted using a model that is trained using a speaker identification or speaker verification task. The embeddings are therefore expected to express characteristics of the audio that are useful in discriminating one speaker from another. It does not consider the locations or movements of the speakers.

In order to use location information, it is proposed in \cite{wong2021} to use a similarity score that is a KL-divergence between SSL vectors, within a HMM diarisation framework. This paper proposes as a baseline, that this score can also be used with AHC, by computing the SSL contribution to the affinity as
\begin{equation}
\mathcal{A}_\text{KL}\left(n,m\right)=\frac{1}{2}\left(\mathbf{s}_n\cdot\frac{\log\mathbf{s}_m}{\log\mathbf{s}_n}+\mathbf{s}_m\cdot\frac{\log\mathbf{s}_n}{\log\mathbf{s}_m}\right),
\label{eq:affinity_kl}
\end{equation}
where $\mathbf{s}_n$ is the SSL centroid of cluster $n$. Unlike in \cite{wong2021}, the symmetric KL-divergence is used here. However, this affinity averages the SSL vectors over time when computing the centroid, and may therefore not explicitly model the movement of the speakers.

This paper proposes to model speaker movements within AHC by computing an affinity based on the log-likelihood ratio between merged clusters and separated clusters, where the log-likelihood is computed using the Kalman filter,
\begin{align}
\mathcal{A}_\text{track}\!\left(n,m\right)\!=\!\frac{1}{\breve{T}_n+\breve{T}_m}\!\!&\left[\log p\left(\mathbf{X}_{t_n^\text{start}:t_n^\text{end}},\mathbf{X}_{t_m^\text{start}:t_m^\text{end}}\right)\right.\label{eq:affinity_llr}\\
&-\left.\log p\left(\mathbf{X}_{t_n^\text{start}:t_n^\text{end}}\right)-\log p\left(\mathbf{X}_{t_m^\text{start}:t_m^\text{end}}\right)\right]\!\!,\notag
\end{align}
where $\breve{T}_n$ is the number of frames that have a DOA or SSL observation in cluster $n$, and $t_n^\text{start}$ and $t_n^\text{end}$ represent the first and last frame indexes of cluster $n$ respectively. The normalisation by the number of frames with observations is necessary, as the log-likelihood from \eqref{eq:conditional_loglikelihood} scales linearly with the number of frames with observations. In the future, it may be interesting to explore BIC equivalents for this affinity, to take into account the model complexity.

The affinities of $\mathcal{A}_\text{KL}$ and $\mathcal{A}_\text{track}$ express the cluster similarity based on the speakers' angular locations, while $\mathcal{A}_\text{speaker}$ expresses the cluster similarity based on the characteristics of the audio that are useful in discriminating one speaker from another. The location affinities alone may not be sufficient for the clustering task, as it is possible that multiple speakers may overlap in their angular locations, or even in their movements. However, an affinity based on location may be complementary to one based on the speaker's acoustic discriminative characteristics. Therefore, it may be useful to interpolate together $\mathcal{A}_\text{speaker}$ and either $\mathcal{A}_\text{KL}$ or $\mathcal{A}_\text{track}$.

\section{Meeting transcription setup}

A rich meeting transcription task was used to evaluate the proposed approach. The setup followed that described in \cite{yoshioka2019b,wong2021}. Audio from a microphone array was beamformed and separated into multiple channels, where it was assumed that there were no concurrent speakers within each channel. On each channel, voice activity detection and speech recognition were run. The non-silence segments were then further split into segments with speaker purity, using speaker change detection at word boundaries. Each of these segments might contain one or more words. Diarisation was then run to cluster and tag the resulting segments from all channels together. For each segment, a speaker embedding was extracted using the model described in \cite{zhou2021}. DOA and SSL features were also extracted, as is described in Section \ref{sec:observation_feature}. AHC was used to cluster the segments from the same speaker together, using a combination of one or more of the affinity measures described in Section \ref{sec:ahc}. Finally, the Hungarian algorithm was used to tag the clusters, by finding an optimal mapping between the clusters and the enrolled speakers, based on the similarity of the speaker embeddings. The AHC affinity of $\mathcal{A}_\text{KL}$ in \eqref{eq:affinity_kl} was computed using a single SSL vector per cluster that represents the centroid, while $\mathcal{A}_\text{track}$ in \eqref{eq:affinity_llr} used a sequence of DOA or SSL features with a duration and shift of 0.4s.

\begin{table}[t]
\centering
\caption{Comparison of Kalman filter location feature types}
\label{tab:feature}
\begin{tabular}{c|c}
\hline
Feature type&\emph{dev} speaker-attributed WER (\%)\\
\hline\hline
DOA&22.71\\
SSL&22.46\\
\hline
\end{tabular}
\end{table}

\section{Experiments}

Experiments were performed on audio collected from internal Microsoft meetings, lasting up to 1 hour each, with an average of 7 active participants each. The \emph{dev} set comprised 51 meetings totalling 23 hours, while the \emph{eval} set comprised 60 meetings totalling 35 hours. The speaker embeddings had 128 dimensions, while the SSL vectors had 360 dimensions. The AHC stopping criterion and affinity interpolation weights were tuned on the \emph{dev} set using parameter sweeps. The Kalman filter parameters were optimised on the \emph{dev} set using the EM algorithm, as is described in Section \ref{sec:parameter_estimation}. The performance was measured using the speaker-attributed Word Error Rate (WER) \cite{yoshioka2019b}. This first computed the WER separately for each speaker, by comparing the hypothesis to the reference for that speaker, then the WERs were averaged over all speakers. The speaker-attributed WER expresses a combination of the speech recognition and diarisation performances, both of which are important for the rich meeting transcription task.

Table \ref{tab:feature} compares computing the log-likelihood ratio using either DOA or SSL features in the Kalman filter, on the \emph{dev} set. In both cases, the location tracking affinity of $\mathcal{A}_\text{track}$ in \eqref{eq:affinity_llr} was interpolated with the speaker embedding affinity of $\mathcal{A}_\text{speaker}$ in \eqref{eq:affinity_dvec}. The results suggest that using the full SSL vectors as the observed location features in the Kalman filter performs better than using DOA features. Each SSL vector is summarised by a mean angle in \eqref{eq:circular_mean} and a concentration in \eqref{eq:ssl_concentration}. The concentration may contain information about the certainty of the instantaneous location estimation, which is not expressed in the DOA features. This weighs the contribution of each frame to the total sequence log-likelihood. The remaining experiment used the SSL features in the Kalman filter.

\begin{table}[t]
\centering
\caption{Usefulness of modelling movement for diarisation}
\label{tab:moving}
\begin{tabular}{c|l|ccc}
\hline
&&\multicolumn{3}{c}{Speaker-attributed WER (\%)}\\
Test set&Affinity&stationary&moving&average\\
\hline\hline
\multirow{3}{*}{\emph{dev}}&$\mathcal{A}_\text{speaker}$&22.61&27.19&25.42\\
&$\mathcal{A}_\text{speaker}+\mathcal{A}_\text{KL}$&22.05&27.55&25.43\\
&$\mathcal{A}_\text{speaker}+\mathcal{A}_\text{track}$&21.10&23.32&22.46\\
\hline
\multirow{3}{*}{\emph{eval}}&$\mathcal{A}_\text{speaker}$&25.32&20.39&23.65\\
&$\mathcal{A}_\text{speaker}+\mathcal{A}_\text{KL}$&24.73&19.83&23.06\\
&$\mathcal{A}_\text{speaker}+\mathcal{A}_\text{track}$&23.65&20.15&22.40\\
\hline
\end{tabular}
\end{table}

The use of the Kalman filter log-likelihood ratio, $\mathcal{A}_\text{track}$ in \eqref{eq:affinity_llr}, within the affinity for AHC can be compared against the baselines of using speaker embeddings with $\mathcal{A}_\text{speaker}$ in \eqref{eq:affinity_dvec} and the KL-divergence based instantaneous location affinity of $\mathcal{A}_\text{KL}$ in \eqref{eq:affinity_kl}. The meetings were categorised into those with and without moving speakers. A meeting was considered to contain moving speakers if for at least one speaker in the meeting, it was possible to find two disjoint angular arcs of at least $\frac{\pi}{6}$ radians each, where that speaker spent at least 30s of active speech in each of the two angular regions not covered by these two arcs, based on manually transcribed location information from video data. The performances of the various AHC affinities on the stationary and moving meetings are shown in Table \ref{tab:moving}. Interpolating $\mathcal{A}_\text{KL}$ with $\mathcal{A}_\text{speaker}$ improves the performance for both stationary and moving meetings on the \emph{eval} set, and the stationary meetings on the \emph{dev} set, compared against using $\mathcal{A}_\text{speaker}$ alone. This agrees with \cite{wong2021} in suggesting that location information may be complementary to the speaker embeddings for the clustering task. Initial tests suggested that it may be difficult to robustly tune the AHC stopping criterion and the affinity interpolation weights between $\mathcal{A}_\text{speaker}$ and $\mathcal{A}_\text{KL}$, as the cosine similarity in $\mathcal{A}_\text{speaker}$ has a dynamic range between -1 and 1, while that for the KL-divergence in $\mathcal{A}_\text{KL}$ is between $-\infty$ and 0. The Kalman filter log-likelihood ratio in $\mathcal{A}_\text{track}$ outperforms $\mathcal{A}_\text{KL}$ on the moving meetings in the \emph{dev} set, but not the \emph{eval} set. This may again suggest that it can be difficult to robustly tune the hyper-parameters for the interpolated affinity to generalise well to new data, since $\mathcal{A}_\text{speaker}$ and $\mathcal{A}_\text{track}$ again have different dynamic ranges. Despite this, $\mathcal{A}_\text{track}$ consistently yields improvements for the moving meeting compared to using $\mathcal{A}_\text{speaker}$ alone. In both datasets, $\mathcal{A}_\text{track}$ is also able to yield consistent improvements over $\mathcal{A}_\text{KL}$ when averaged over all meetings, thereby suggesting that there may be a benefit to explicitly model speaker movements when performing diarisation.

\section{Conclusion}

This paper has presented an approach to explicitly model the spatial movements of speakers while performing diarisation. The movements are modelled through location tracking, using a Kalman filter with von Mises density functions as the transition and emission likelihoods. This Kalman filter is used to compute log-likelihood ratios between different cluster merging hypotheses in AHC. The results suggest that explicitly modelling the movements of speakers may provide information that is complementary to the speaker embeddings for the diarsation task.

\bibliographystyle{IEEEbib}
\bibliography{strings,refs}

\end{document}